\documentclass{llncs}
\usepackage{mathptmx}
\usepackage{times}
\usepackage{helvet}
\usepackage{courier}
\usepackage{multicol}
\usepackage{microtype}

\usepackage{multirow}
\usepackage{xspace}
\usepackage{paralist}

\newif\ifanonymous\anonymousfalse
\newif\ifextended\extendedfalse
\newif\ifappendix\appendixfalse
\newif\ifdotikz\dotikztrue
\newif\ifdraft\drafttrue
\dotikzfalse
\draftfalse

\usepackage{epsfig}
\usepackage{longtable}
\usepackage{multirow}
\usepackage{color}
\usepackage{amsfonts,url}
\usepackage{amsmath}

\usepackage{xspace}
\usepackage{paralist}

\ifdraft
  \usepackage{todonotes}
  \newcommand{\comment}[1]{\todo[inline,color=green!25]{#1}}
\else
  \newcommand{\comment}[1]{}
\fi

\graphicspath{{figures/}}

\ifdotikz
  \usepackage{tikz}
  \usetikzlibrary{shapes,arrows,backgrounds,%
  matrix,patterns,arrows,decorations.pathmorphing,decorations.pathreplacing,%
  positioning,fit,calc,decorations.text,shadows%
  }
\else
  \long\def\beginpgfgraphicnamed#1#2\endpgfgraphicnamed{\includegraphics{#1}}
\fi

\newcommand{\leanparagraph}[1]{\smallskip\par\noindent{\bfseries #1.}\ }

\newcommand{\bi}[0]{\begin{itemize}}
\newcommand{\ei}[0]{\end{itemize}}
\newcommand{\mi}[1]{\ensuremath{\mathit{#1}}}

\def\gringo{{\sc Gringo}\xspace}
\def\clasp{{\sc Clasp}\xspace}

\def\python{\textrm{Python}\xspace}

\newcommand{\hex}{\textsc{hex}\xspace}
\newcommand{\dlvhex}{\textrm{dlvhex}\xspace}

\def\ccalc{{\sc CCalc}\xspace}

\def\ba{\begin{array}}
\def\ea{\end{array}}
\def\beq{\begin{equation}}
\def\eeq#1{\label{#1}\end{equation}}

\def\eqs{\,{=}\,}

\def\timess{\,{\times}\,}

\newcommand{\mypre}{\textsc{Pre}\xspace}
\newcommand{\myint}{\textsc{Int}\xspace}
\newcommand{\mypost}{\textsc{Post}\xspace}
\newcommand{\myfilt}{\textsc{Filt}\xspace}
\newcommand{\myrepl}{\textsc{Repl}\xspace}
\newcommand{\myLpre}{\ensuremath{L_{\mypre}}\xspace}
\newcommand{\myLil}{\ensuremath{L_{\myint}}\xspace}
\newcommand{\myLpost}{\ensuremath{L_{\mypost}}\xspace}
\newcommand{\myfirst}{\textsc{first}\xspace}
\newcommand{\myall}{\textsc{all}\xspace}

\newcommand\myTikzDataflow{%
  \beginpgfgraphicnamed{dataflow}%
  \small%
  \begin{tikzpicture}[inner sep=0.2em,outer sep=0em,
    op/.style={draw,shape=rectangle,minimum width=7.0em,minimum height=1.8em},
    llr/.style={draw,shape=cylinder,minimum width=1.9em,shape aspect=0.5,minimum height=3.2em},
    lar/.style={draw,-latex'}
    ]%
  \node[op] (pre) {Precomputation};
  \node[op,below=2em of pre] (hex)
    {\begin{tabular}{@{}c@{}} Planning\\ (ASP Solver) \end{tabular}};
  \node[op,below=2em of hex] (post) {Postcheck};
  \node[llr,right=0ex of pre] {\myLpre};
  \node[llr,right=0ex of hex] {\myLil};
  \node[llr,right=0ex of post] {\myLpost};

  \draw[lar] (pre.south) -- node[left] {$H^\mi{pre}$\,} (hex.north);
  \draw[lar] (hex.south) -- node[left] {candidate solution\,} (post.north);
  \draw[lar] (post.south) -- node[left,near end] {feasible solution\,} ($(post.south)-(0,2em)$);

  \draw[lar] ($(pre.north)+(-4em,1.0em)$) -| (pre.north);
  \node at ($(pre.north)+(-5em,0.8em)$) {$H$};

  \draw[lar] ($(post.south)+(2em,0)$) --
    ++(0,-1em) -- ++(6em,0) -- ++(0,8.7em) -- ++(-6em,0) -- ($(hex.north)+(2em,0)$);
  \node at ($(hex.east)+(7.9em,0.1em)$) {$H^+$ constraints};
  \node at ($(hex.west)-(8em,0em)$) {~};
  \end{tikzpicture}
  \endpgfgraphicnamed
}

\newcommand{\myframeworkfigure}[1]{
\begin{figure}[#1]
  \begin{center}
  \myTikzDataflow
  \end{center}
  \caption{Components and data flow.} %
  \label{fig:flows}
  \vspace{-1.5em}
\end{figure}
}

\newcommand{\hphup}{\hphantom{\,(0)}}
\newcommand{\mylr}{}%
\newcommand{\myna}{0}

\newcommand{\myllrange}{\ensuremath{L_{\mi{leg}}}\xspace}
\newcommand{\myllbalance}{\ensuremath{L_{\mi{bal}}}\xspace}
\newcommand{\myrmrobot}{\ensuremath{L_{\mi{rob}}}\xspace}
\newcommand{\myrmpayload}{\ensuremath{L_{\mi{pay}}}\xspace}

\newcommand\myresultstableefftabular{
\centering{%
\small
\begin{tabular}{@{~}c@{~}@{~\;}r@{\;~}@{~\;}r@{~\;}@{\;}r@{~}@{~}r@{\;}}
\hline
Integration\mylr&
    \multicolumn{2}{c@{\;}@{\;}}{Overall Time} &
        \multicolumn{2}{@{\;}c@{\;}}{Low-Level Reasoning} \\
Method & %
    ~~\myfirst~~ & ~~\myall~~~ &
        ~time \myall & count \myall \\
&
    ~~~~sec~~~~ & ~~~~sec~~~~ & sec & \# \\
\hline
\multicolumn{5}{c}{Robotic Manipulation (10 instances)\mylr} \\
\hline
\textsc{Filt} & 1716\,[2]  & 1877\,[2]  &  39 &   724\mylr \\
\textsc{Repl} & 2007\,[2]  & 3242\,[3]  &   7 &   139      \\
\textsc{Pre}  &  888\hphup &  974\hphup & 238 & 29282      \\
\textsc{Int}  & 1007\hphup & 1086\hphup &   0 &   467      \\
\hline
\multicolumn{5}{c}{Legged Locomotion (averages over 20 instances)\mylr} \\
\hline
\textsc{Filt} & 2434\,[5]  & 3091\,[8]   & 1139 &  35888\mylr \\
\textsc{Repl} & 1345\hphup & 4192\,[9]   &   12 &    458      \\
\textsc{Int}  &   80\hphup &  133\hphup  &   21 & 171109      \\
\multirow{4}{*}{%
    \begin{tabular}{@{}l@{~}l@{}}
    \myllrange: & \myllbalance: \\
    \textsc{Pre} & \textsc{Filt} \\
    \textsc{Pre} & \textsc{Repl} \\
    \textsc{Pre} & \textsc{Int}
    \end{tabular}} & & & & \\
          & 2395\,[6]  & 3046\,[8]  & 1272 & 39354 \\
          & 1160\hphup & 4142\,[9]  &    9 &   324 \\
          &   65\hphup &  107\hphup &   23 & 50677 \\
\hline
\multicolumn{5}{@{\,}l@{\,}}{Numbers in square brackets count timeouts for \myfirst resp.~\myall.\mylr}
\end{tabular}%
}%
}

\newcommand{\myresultstableeff}[1]{
\begin{table}[#1]%
\caption{Efficiency Comparison}%
\label{tblEff}%
\myresultstableefftabular
\end{table}
}

\newcommand\myresultstablequaltabular{
\centering{%
\small
\begin{tabular}{@{\;}c@{\;}@{\;}r@{\;}@{\;}r@{\;}@{\;}r@{\;}@{\;}r@{\;}}
\hline
Integration\mylr&
  \multicolumn{2}{@{\;}c@{\;}@{\;}}{\!Infeasible Candidates} &
  \multicolumn{1}{@{\;}c@{\;}@{\;}}{\!Plans found} &
  Feasible Plans \\
Method&
  \multicolumn{1}{@{}c@{\;}@{\;}}{~~~\myfirst~~~} &
  \multicolumn{1}{@{\;}c@{\;}@{\;}}{\myall} &
  \multicolumn{1}{@{\;}c@{\;}@{\;}}{\myall} &
  \multicolumn{1}{@{}c@{\;}}{\myall}  \\
& \multicolumn{1}{@{\;}c@{\;}@{\;}}{\#} &
    \multicolumn{1}{@{\;}c@{\;}@{\;}}{\#} &
  \multicolumn{1}{@{\;}c@{\;}@{\;}}{\#} &
    \multicolumn{1}{@{}c@{\;}}{\%}  \\
\hline
\multicolumn{5}{c}{Robotic Manipulation (averages over 10 instances)\mylr} \\
\hline
\textsc{Filt} &   586 & 11787 & 622 & ${<}$0.1\mylr \\ %
\textsc{Repl} &    11 &    38 & 621 &     94.2      \\ %
\textsc{Pre}  & \myna & \myna & 652 &    100.0      \\ %
\textsc{Int}  & \myna & \myna & 652 &    100.0      \\ %
\hline
\multicolumn{5}{c}{Legged Locomotion (averages over 20 instances)\mylr} \\
\hline
\textsc{Filt} & 11282 & 35487 &  360 &      1.0\mylr \\ %
\textsc{Repl} &    28 &    68 &  250 &     78.6      \\ %
\textsc{Int}  & \myna & \myna & 1394 &    100.0      \\ %
\multirow{4}{*}{%
    \begin{tabular}{@{}l@{~}l@{}}
    \myllrange: & \myllbalance: \\
    \textsc{Pre} & \textsc{Filt} \\
    \textsc{Pre} & \textsc{Repl} \\
    \textsc{Pre} & \textsc{Int}
    \end{tabular}} & & & & \\
          & 10938 & 39116 &  340 &    0.9 \\ %
          &    31 &    69 &  255 &   78.7 \\ %
          & \myna & \myna & 1394 &  100.0 \\ %
\hline
\end{tabular}%
}%
}

\newcommand{\myresultstablequal}[1]{
\begin{table}[#1]%
\caption{Solution Quality Comparison}%
\label{tblQual}%
\myresultstablequaltabular
\end{table}
}

\newcommand\bcmdtab{\noindent\bgroup\tabcolsep=0pt%
  \begin{tabular}{@{}p{10pc}@{}p{20pc}@{}}}
\newcommand\ecmdtab{\end{tabular}\egroup}

  \title%
        {Levels of Integration between\\ Low-Level Reasoning and Task Planning}

  \author%
         {Esra Erdem, Volkan Patoglu, Peter Sch\"{u}ller}
   \institute{Faculty of Engineering and Natural Sciences, Sabanc{\i} University, \.Istanbul, Turkey\\
          \email{\{esraerdem,vpatoglu,peterschueller\}@sabanciuniv.edu}}

\begin{document}

\label{firstpage}

\maketitle

  \begin{abstract}
We provide a systematic analysis of levels of integration between
discrete high-level reasoning and continuous low-level reasoning to
address hybrid planning problems in robotics.  We identify four
distinct strategies for such an integration: (i) low-level checks
are done for all possible cases in advance and then this information
is used during plan generation, (ii) low-level checks are done
exactly when they are needed during the search for a plan, (iii)
first all plans are computed and then infeasible ones are filtered,
and (iv) by means of replanning, after finding a plan, low-level
checks identify whether it is infeasible or not; if it is
infeasible, a new plan is computed considering the results of
previous low-level checks. We perform experiments on hybrid planning
problems in robotic manipulation and legged locomotion domains considering
these four methods of integration, as well as some of their
combinations.  We analyze the usefulness of levels of integration in
these domains, both from the point of view of computational
efficiency (in time and space) and from the point of view of plan
quality relative to its feasibility. We discuss advantages and
disadvantages of each strategy in the light of experimental results
and provide some guidelines on choosing proper strategies for a
given domain.
  \end{abstract}


\section{Introduction}

Successful deployment of robotic assistants in our society requires
these systems to deal with high complexity and wide variability of
their surroundings to perform typical everyday tasks robustly and
without sacrificing safety. Consequently, there exists a pressing
need to furnish these robotic systems not only with discrete high-level
reasoning  (e.g., task planning, diagnostic reasoning) and
continuous low-level reasoning (e.g., trajectory planning, deadline
and stability enforcement) capabilities, but also their tight integration
resulting in hybrid planning.

Motivated by the importance of hybrid planning, recently there have
been some studies on integrating discrete task planning and
continuous motion planning. These studies can be grouped into two,
where integration is done at the search level or at the
representation level.
For instance,
\cite{Asymov_book,Hauser2009,KaelblingL11,Plaku2010,Wolfe2010,Plaku12}
take advantage of a forward-search task planner to incrementally
build a task plan, while checking its kinematic/geometric
feasibility at each step by a motion planner; all these approaches
use different methods to utilize the information from the task-level
to guide and narrow the search in the configuration space. By this
way, the task planner helps focus the search process during motion
planning. Each one of these approaches presents a specialized
combination of task and motion planning at the search level, and
does not consider a general interface between task and motion
planning.

On the other hand,
\cite{cal09,HertleDKN12,2011_combining_high_level_causal_reasoning_with_low_level_geometric_reasoning_and_motion_planning_for_robotic_manipulation,erdemAP12}
integrate task and motion planning by considering a general
interface between them, using ``external predicates/functions'',
which are predicates/functions that are computed by an external
mechanism, e.g., by a C++ program. The idea is to use external
predicates/functions in the representation of actions, e.g., for
checking the feasibility of a primitive action by a motion planner.
So, instead of guiding the task planner at the {\em search level} by
manipulating its search algorithm directly, the motion planner
guides the task planner at the {\em representation level} by means
of external predicates/functions.
\cite{cal09,2011_combining_high_level_causal_reasoning_with_low_level_geometric_reasoning_and_motion_planning_for_robotic_manipulation}
apply this approach in the action description language ${\cal C}+$
\cite{giu04} using the causal reasoner \ccalc~\cite{mcc97a};
\cite{erdemAP12} applies it in Answer Set Programming
(ASP)~\cite{lifschitz08,BrewkaET11} using the ASP
solver~\clasp~\cite{gebserKNS07}; \cite{HertleDKN12} extends the
planning domain description language PDDL \cite{pddl03} to support
external predicates/functions (called semantic attachments) and
modifies the planner FF \cite{ff01}
accordingly.

In these approaches, integration of task and motion planning is
achieved at various levels. For instance,
\cite{2011_combining_high_level_causal_reasoning_with_low_level_geometric_reasoning_and_motion_planning_for_robotic_manipulation,erdemAP12}
do not delegate all sorts of feasibility checks to external
predicates as in \cite{cal09,HertleDKN12}, but implements only some
of the feasibility checks (e.g., checking collisions of robots with
each other and with other objects, but not collisions of objects
with each other) as external predicates and use these external
predicates in action descriptions to guide task planning. For a
tighter integration, feasibility of task plans is checked by a
dynamic simulator; in case of infeasible plans, the planning problem
is modified with respect to the causes of infeasibilities, and the
task planner is asked to find another plan.

In this paper, our goal is to better understand how much of
integration between high-level reasoning and continuous low-level
reasoning is useful, and for what sort of robotic applications. For
that, we consider integration at the representation level, since
this approach allows a modular integration via an interface, external
predicates/functions, which provides some flexibility of embedding
continuous low-level reasoning into high-level reasoning at various
levels. Such a flexible framework allowing a modular integration
is important for a systematic analysis of levels of integration.

We identify four distinct strategies to integrate a set of
continuous feasibility checks into high-level reasoning, grouped
into two: \emph{directly integrating} low-level checks into
high-level reasoning while a feasible plan is being generated, and
generating candidate plans and then \emph{post-checking} the
feasibility of these candidate solutions with respect to the
low-level checks. For direct integration we investigate two methods
of integration: (i) low-level checks are done for all possible cases
in advance and then this information is used during plan generation,
(ii) low-level checks are done when they are needed during the
search for a plan. For post-checking we look at two methods of
integration: (iii) all plans are computed and then infeasible ones
are filtered, (iv) by means of replanning, after finding a plan,
low-level checks identify whether it is infeasible or not; if it is
infeasible, a new plan is computed considering the results of
previous low-level checks. We consider these four methods of
integration, as well as some of their combinations; for instance,
some geometric reasoning can be integrated within search as needed,
whereas some temporal reasoning is utilized only after a plan is
computed in a replanning loop. Considering each method and some of
their combinations provide us different levels of integration.

To investigate the usefulness of these levels of integration at
representation level, we consider 1)  the expressive formalism of
\hex\ programs for describing actions and the efficient \hex solver
\dlvhex\ to compute plans, and 2) the expressive formalism of ASP
programs for describing actions and the efficient ASP solver \clasp\
to compute plans.
Unlike the formalisms and solvers used in other
approaches~\cite{cal09,HertleDKN12,2011_combining_high_level_causal_reasoning_with_low_level_geometric_reasoning_and_motion_planning_for_robotic_manipulation,erdemAP12},
that study integration at representation level,
\hex~\cite{2005_a_uniform_integration_of_higher_order_reasoning_and_external_evaluations_in_answer_set_programming}
and \dlvhex~\cite{dlvhex} allow external predicates/functions to
take relations (e.g., a fluent describing locations of all objects)
as input without having to explicitly enumerate the objects in the
domain. Other formalisms and solvers allow external
predicates/functions to take a limited number of objects and/or
object variables as input only, and thus they do not allow embedding
all continuous feasibility checks in the action descriptions. In
that sense, the use of \hex programs with \dlvhex, along with the
ASP programs with \clasp enriches the extent of our experiments.

We perform experiments on planning problems in a robotic manipulation domain
(like in~\cite{2011_combining_high_level_causal_reasoning_with_low_level_geometric_reasoning_and_motion_planning_for_robotic_manipulation})
and in a legged locomotion domain (like
in~\cite{2005_multi_step_motion_planning_for_free_climbing_robots,2004_free_climbing_with_a_multi_use_robot}).
Robotic manipulation domain involves 3D collision checks and inverse kinematics,
whereas legged locomotion involves
stability and reachability checks. We analyze the usefulness of
levels of integration in these domains, both from the point of view
of computational efficiency (in time and space) and from the point
of view of plan quality relative to its feasibility.

\section{Levels of Integration}

Assume that we have a task planning problem instance $H$ (consisting
of an initial state $S_0$, goal conditions, and action descriptions)
in a robotics domain, represented in some logic-based formalism. A
history of a plan $\langle A_0,\dots,A_{n-1}\rangle$ from the given
initial state $S_0$ to a goal state $S_n$ computed for $H$ consists
of a sequence of transitions between states: $\langle
S_0,A_0,S_1,A_1,\dots,S_{n-1},A_{n-1},S_n\rangle$.
A {\em low-level continuous reasoning module} gets as input,
a part of a plan history
computed for $H$ and returns whether this part of the plan history is
feasible or not with respect to some geometric, dynamic or temporal
reasoning.

For example, if the position of a robot at step $t$ is represented
as $\mi{robot\_at}(x,y,t)$ and the robot's action of moving to
another location $(x',y')$ at step $t$ is represented  as
$\mi{move\_to}(x',y',t)$, then a motion planner could be used to
verify feasibility of the movement $\langle \mi{robot\_at}(x,y,t),$
$\mi{move\_to}(x',y',t), \mi{robot\_at}(x',y',t+1) \rangle$.
If duration of this action is represented as well, e.g., as
$\mi{move\_to}(x,y,\mi{duration},t)$, then the low-level module can
find an estimate of the duration of this movement relative to the
trajectory computed by a motion planner, and it can determine the
feasibility of the movement $\langle \mi{robot\_at}(x,y,t),
\mi{move\_to}(x',y',t), \mi{robot\_at}(x',y',t+1) \rangle$ by
comparing this estimate with $\mi{duration}$.

Let $L$ denote a low-level reasoning module that can be used
for the feasibility checks of plans for a planning problem instance
$H$.
We consider four different methods of utilizing $L$ for computing
feasible plans for $H$, grouped into two:
\emph{directly integrating} reasoning $L$ into $H$, and
\emph{post-checking} candidate solutions of $H$ using $L$.

For \emph{directly integrating} low-level reasoning into plan
generation,  we propose the following two levels of integration:
\begin{itemize}
\item
{\bf\mypre} --
\emph{Precomputation} %
We perform all possible feasibility checks of $L$ that can be
required by $H$, in advance.
For each failed check, we identify actions that cause the
failure, and then add a constraint to the action descriptions in $H$
ensuring that these actions do not occur in a plan computed for $H$.
We then try to find a plan for the augmented planning
problem instance $H^{\mi{pre}}$.
Clearly, every plan obtained with this method satisfies all
low-level checks.

\item
{\bf\myint} --
\emph{Interleaved Computation} %
We do not precompute %
but we interleave low-level checks with high-level reasoning in the search of a plan:
for each action considered during the search,
the necessary low-level checks are immediately performed
to find out whether including this action will lead to an infeasible plan.
An action is included in the plan only if it is feasible.
The results of feasibility checks of actions can be stored
not to consider infeasible actions repeatedly in the search of a plan.
Plans generated by interleaved computation satisfy all low-level checks.
\end{itemize}
Let us denote by $\myLpre$ and $\myLil$ the low-level checks directly integrated into plan generation,
with respect to \mypre and \myint, respectively.

Alternatively, we can integrate low-level checks $L$ with $H$, by means of
\emph{post-checking} candidate solutions of $H$ relative to $L$.
We propose the following two methods to perform post-checks on solution candidates:
\begin{itemize}
\item
{\bf\myfilt} --
\emph{Filtering}:
We generate all plan candidates for $H$.
For each low-level check in $L$,
we check feasibility of each plan candidate
and discard all infeasible candidates.

\item
{\bf\myrepl} --
\emph{Replanning}:
We generate a plan candidate for $H$.
For each low-level check in $L$,
we check feasibility of the plan candidate.
Whenever a low-level check fails,
we identify the actions that cause the failure, and
then add a constraint to $H$ ensuring that these actions do not occur in a plan computed for $H$.
We generate a plan candidate for the updated planning problem instance $H^+$
and do the feasibility checks. We continue with generation of plan candidates and low-level
checks until we find a feasible plan, or find out that such a feasible plan does not exist.
\end{itemize}
Let us denote by $\myLpost$ the low-level checks done after plan generation,
with respect to \myfilt or \myrepl.

Figure~\ref{fig:flows} shows the hybrid planning framework we use in this paper
to compare different levels of integration, and combinations thereof,
on robotics planning scenarios. In particular, Fig.~\ref{fig:flows} depicts computational components:
Precomputation extends the problem instance $H$ using a low-level reasoning module $\myLpre$,
Planning integrates a low-level reasoning module $\myLil$ into its search
for a plan candidate for the problem instance $H^{\mi{pre}}$ generated by Precomputation.
Postcheck uses a low-level module $\myLpost$
to verify solution candidates (using \myfilt or \myrepl)
and to potentially add constraints $H^+$ to the input of Planning.

In our systematic analysis of levels of integration, we do consider this hybrid framework
by disabling some of its components. For instance, to analyze the usefulness of \mypre,
we disable the other integrations (i.e., $\myLil=\myLpost=\emptyset$); to analyze the usefulness of
a combination of \mypre and \myfilt, we disable other integrations (i.e., $\myLil=\emptyset$).

\section{Methodology}
We investigate the usefulness of levels of integration as described
above, considering two orthogonal properties: solution quality and
planning efficiency.
We quantify these properties as follows.

\myframeworkfigure{t!}

\leanparagraph{Solution Quality}
If some low-level module $L$ is not integrated into the planning process,
some plan candidates will be infeasible due to failed low-level checks of $L$.
We quantify solution quality by measuring the number of feasible and infeasible plan candidates
generated by the search for a plan.
This way we obtain a measure that shows how \emph{relevant} a given low-level check is for plan feasibility.
Note that with the \myfilt approach an infeasible plan candidate
simply causes a new plan to be generated,
while with \myrepl an infeasible plan candidate causes computation of additional constraints,
and a restart of the plan search.

Tightly connected to the number of feasible and infeasible solution candidates
is the number of low-level checks that is performed until finding the first feasible plan,
and until finding all feasible plans.

\leanparagraph{Planning Efficiency}
We quantify planning efficiency by measuring the time required
to obtain the first feasible plan,
and the time to enumerate all feasible plans.
(Note that this includes proving that no further plan exists.)

Independent from the number of low-level checks, the duration of
these external computations can dominate the overall planning cost,
or it can be negligible.
Therefore we measure %
not only the number of computations of low-level modules
but also the time spent in these computations.

\section{Domains and Experimental Setup}

For our empirical evaluation %
we use the Robotic Manipulation and the Legged Locomotion domains.
Both require hybrid planning.
We next give an overview of the domains, their characteristics,
and scenarios we used.

\leanparagraph{Robotic Manipulation}
We consider a cooperative robotic manipulation problem, as in~%
\cite{2011_combining_high_level_causal_reasoning_with_low_level_geometric_reasoning_and_motion_planning_for_robotic_manipulation},
where two robots arrange elongated objects in a space that contains obstacles.
The manipulated objects can only be carried cooperatively by both robots,
objects must not collide with each other or the environment,
similarly robots must not collide with each other.

A large part of collision checks between objects
can already be realized in the high-level representation,
however certain checks require usage of geometric models.
Collision-freeness between robots for particular collaborative
actions can only be determined using low-level geometric reasoning
and is not represented in the world model.

Therefore we use two low-level reasoning components to check collision-freeness:
the \myrmrobot module checks collisions between the robotic arms,
and \myrmpayload checks collisions between an object and its environment.
We experiment with 10 instances (over a $11{\times}11$ grid) that
require plans of upto 20 (average 9.2) steps, and involving up to 58
(average 25.1) actions.

\leanparagraph{Legged Locomotion}
In the Legged Locomotion domain,
a robot with high degrees of freedom must find a plan
for placing its legs and moving its center of mass (CM) in order to move from one location to another one.

For the purpose of studying integration of geometric reasoning with
high-level task planning, we created a planning formulation for a
four-legged robot that moves on a $10\timess10$ grid.
Some grid locations are occupied and must not be used by the robot.
Starting from a given initial configuration, the robot must reach a
specified goal location where all legs are in contact with the
ground.

As legged robots have high degrees of freedom, legged locomotion
planning deals with planning in a high-dimensional space.
We use a planning problem that is of similar complexity as has been investigated
in climbing~\cite{2004_free_climbing_with_a_multi_use_robot}
and walking~\cite{2008_motion_planning_for_legged_robots_on_varied_terrain} robots.
We also require a feasibility check of leg placement actions.
We allow concurrent actions, i.e., moving the center of mass while detaching a leg from the ground,
if this does not cause the robot to lose its balance.

We use a low-level reasoning component that determines whether the
robot is in a balanced stable equilibrium (\myllbalance), given its
leg positions and the position of its CM.
We realize this check by computing the support polygon of legs that are currently connected to the ground,
and by checking if CM is within that polygon.
For these checks we use the {\tt boost::geometry} library to compute a convex hull of all leg positions,
and then check whether CM is located within that convex hull.

A second low-level module determines if leg positions are realistic wrt.\ the position of CM,
i.e., if every leg can reach the position where it is supposed to touch to the ground.
This check (\myllrange) is realized as a distance computation between coordinates of legs and CM.

\subsection{Domain Characteristics and Notable Differences}
\label{sec:characteristics}

The domains we experiment with exhibit various differences in their
characteristics, and such a variety allows us to get practically
more relevant results.
The most important differences between these two domains are as
follows.

\leanparagraph{Complexity of low-level reasoning}
In Legged Locomotion we use a C++ geometric library
to perform basic geometric operations
which are sufficient for computing check results.

In Robotic Manipulation,
object collision checking \myrmpayload operates on 3D models of objects and environments,
and \myrmrobot additionally requires inverse kinematics to determine
the joint configuration of each robot reaching a certain point
before performing collision checks between arms.

Hence, in Legged Locomotion, each low-level check
requires less time and memory than in Robotic Manipulation.

\leanparagraph{Information relevant for low-level reasoning}
In Legged Locomotion, we consider problem instances  over a $10{\times}10$ grid.
\myllrange is a check over two coordinates,
therefore there are $10^4$ possible \myllrange checks.
The balance check \myllbalance is a totally different situation: we
have an input of four leg coordinates and one CM coordinate,
therefore, there are $10^{10}$ possible \myllbalance checks.
Such a large number of checks makes precomputation
infeasible.

In Robotic Manipulation, both low-level checks are over coordinate
pairs on a $11{\times}11$ grid; therefore, there are  $11^4\cdot 2
\eqs 29282$ low-level checks.

Based on the number of low-level checks, precomputation for Legged
Locomotion seems feasible for only one of the two low-level modules
(\myllrange), while for Robotic Manipulation we can apply
precomputation for both low-level computations.
Indeed, precomputation for Legged Locomotion can be done in less
than 1 second, and for Robotic Manipulation in 238
seconds.\footnote{All experiments were performed on a Linux %
server with 32 2.4GHz Intel\textsuperscript{\textregistered} E5-2665
CPU cores and 64GB memory.}

\section{Experimental Results}
\label{sec:experiment}

We applied different integration methods
to 20 Legged Locomotion and 10 Robotic Manipulation instances of varying size and difficulty.

Tables~\ref{tblEff} and~\ref{tblQual} present results for
\begin{itemize}
\item {\bf\myfirst}:
  obtaining the first feasible plan, and for
\item {\bf\myall}:
  obtaining all (maximum 10000) feasible plans.
\end{itemize}

In our experiments, we use a timeout of 2 hours (7200 seconds) after
which we stop computation and take measurements until that moment.

We also limit the number of enumerated plans to 10000 plans.
The measurements for enumerating up to 10000 plans
reveal information about solution quality and
provides a more complete picture of the behavior of each method:
one method might find a feasible solutions very fast by chance,
whereas finding many or all solutions fast by chance is unlikely.

\leanparagraph{Time Measurements}
Table~\ref{tblEff} shows measurements about planning efficiency
and time spent in low-level reasoning.

Firstly, it is clear that \mypre and \myint --- the direct
integration methods --- outperform \myfilt and \myrepl --- the
post-checking methods: for Robotic Manipulation, only \mypre and
\myint are able to enumerate all solutions within the given time
limit; for Legged Locomotion, only \myint and the \mypre/\myint
combination enumerates all solutions.

Comparing the times required by \mypre and \myint, we see that
\mypre is more efficient for Robotic Manipulation (888 sec vs 1007
sec on average), which is mainly due to efficient precomputation
(see below).

Even though \mypre performs better than \myint,
it spends more time in low-level reasoning,
hence high-level reasoning is faster there;
we can explain this by a more constrained search space
(low-level check results constraint the search).

After \mypre and \myint, the next best choice is \myrepl: it finds
solutions to 8 out of 10 instances in Robotic Manipulation, and it
finds solutions to all instances for Legged Locomotion, whereas
\myfilt has the same number of timeouts in the Manipulation domain
and 5 timeouts for Legged Locomotion.
In addition to that, we can see that \myrepl spends little time in
low-level checks compared to other approaches.
This is because \myrepl performs many restarts of the high-level
planner which causes it to spend a disproportionate amount of time
in high-level planning.
Nevertheless, \myrepl shows its robustness by finding solutions to
all but 2 instances.

Finally, \myfilt fails to find solutions for 7 instances in total
which clearly makes it the worst-performing method.
The time results for Robotic Manipulation suggest that \myfilt may
be a bit faster than \myrepl; this may be an effect of some easy
instances in that domain where replanning spends more time by
reinitialization, than \myfilt spends by iterating over many similar
infeasible solutions.
Therefore, even in that domain, we would not suggest to use \myfilt,
as it might --- by chance, as low-level reasoning cannot give
feedback to high-level reasoning --- fail to find a feasible
solution for a long time.

\myresultstableeff{tbp}

\leanparagraph{Effort of Low-Level Reasoning}
In Robotic Manipulation, while attempting to enumerate all
solutions, \myfilt performs only 724 low-level checks compared to
29282 checks of \mypre.
Similarly, in Legged Locomotion, \myfilt performs 35888 checks and
fails to enumerate all solutions for 8 of 20 instances, while \myint
enumerates all solutions while performing more (171109) low-level
checks.
Note that these numbers (the last column of Table~\ref{tblEff})
indicate \emph{distinct} low-level reasoning tasks
as we cache low-level check results.
These numbers show that \myfilt encounters a small fraction of the
low-level checks that are needed to verify all solutions in \myint.
Caching in fact allows \myfilt to verify much more actions than
\myint (numbers not shown), however the number of \emph{distinct}
checks (numbers shown) is higher in \myint.
We conclude that \myint traverses the solution space much more
efficiently.

In Legged Locomotion, low-level checks depend on a large part of the
candidate plan, so caching is not as effective as in Robotic
Manipulation.
This, together with the fact that in \myfilt the high-level is not
guided by low-level checks, causes the \myfilt approach to spend
more time in low-level reasoning than other approaches.

Note that, to obtain a reasonable comparison between \mypre and the
other approaches, we include times and counts of precomputed
low-level checks in Table~\ref{tblEff} (which explains the large
values for low-level computations in these rows).

\myresultstablequal{tbp}

\leanparagraph{Solution Quality}
Methods \mypre and \myint do not generate infeasible solution
candidates, as they use all low-level checks already in search.

If we compare the number of infeasible solution candidates of
\myfilt and \myrepl in Robotic Manipulation, we observe that \myfilt
generates mainly infeasible solution candidates compared to the
number of feasible solutions (11787 vs 622) while \myrepl creates
only 38 infeasible candidates while enumerating 621 feasible plans.

In Legged Locomotion, the results for \myfilt are similar, however
\myrepl performs a bit worse than in Robotic Manipulation with 250
infeasible candidates compared to 68 feasible solutions.
A possible reason for this difference could be the same reason why
\mypre is not feasible in that domain: there is a large amount of
possible inputs to \myllbalance compared to the other low-level
checks we used.
Due to the large input space, each failed \myllbalance check
constrains the search space only by a small amount, so \myrepl
produces more infeasible solutions than in Robotic Manipulation.

\leanparagraph{Memory Usage}
We measured peak memory usage over the whole runtime of each instance.
Interleaved computation with the \dlvhex solver (columns with \myint)
requires an average of around 2000MB, the maximum stays below 4000MB.
For non-interleaved computations,
\gringo and \clasp were connected with low-level checks using \python scripts.
These approaches require around 400MB of memory with a maximum below 1000MB.

\leanparagraph{Combination of \mypre with other methods}
As shown in the Legged Locomotion experiments,
\mypre can be combined with other approaches.
In our experiments we observe that adding \mypre increases efficiency.

However, \mypre adds a fixed cost to solving because it %
precomputes many points.
Depending on efficiency of low-level computations,
even if there are few possible input combinations to low-level checks %
precomputation might be infeasible.

Dedicated precomputation methods can be more efficient
than just checking for each possible input combination,
e.g., by saving on motion planner initialization.
In our experiments we created such dedicated precomputation methods:
for Robotic Manipulation dedicated precomputation takes 238 seconds in total,
calling individual checks requires 1361 seconds in total.
Without dedicated efficient precomputation,
\mypre performs worse than \myint.

\section{Discussion and Conclusion}
\label{sec:conc}

Our experiments suggest the following conclusions.
If robust and highly complex reasoning is required, and if this
reasoning is done frequently (so that performance gains will become
relevant) then using full interleaved reasoning (\myint) is the only
good option.
\myint has the best performance with respect to run times, and it
can enumerate most solutions compared to other approaches.
The reason is that \myint uses only those low-level checks which are
necessary (they are computed on demand) and therefore does not
overload the solver with redundant information (as \mypre does).
Furthermore, \myint considers failed checks in the search process
and thereby never picks an action where it is known that the action
will violate a low-level check.
This is similar as in the \myrepl approach, but much more efficient
as the integration is much tighter compared to \myrepl.
However, the performance of \myint comes at a price:
\begin{inparaenum}[(a)]
\item
  it requires more memory, and
\item
  it requires a solver that allows for
  interacting with the search process in a tight way,
  usually through an API that has to be used in a sophisticated way
  to be efficient.
\end{inparaenum}

If reasoning operates on a manageable amount of inputs, such that
precomputation is a feasible option, then \mypre is a good choice.
In our Robotic Manipulation experiments, \mypre outperforms all
other methods, which is partially due to our using a dedicated
efficient precomputation tool. %
In Legged Locomotion, combining \mypre with other methods also
increased efficiency.

The \myfilt approach performs the worst, because nothing guides the
search into the direction of a feasible solution; \myfilt is not
robust and enumerates many infeasible solutions.

If both \mypre and \myint are not possible then \myrepl should be
used; this approach does not have the same performance as \myint and
\mypre, however it is a very robust approach as it is guided by its
wrong choices --- we can think of the constraints that are added for
failed low-level checks as the approach `learning from its
mistakes'.
The benchmark results for Legged Locomotion clearly show the
robustness of \myrepl compared to \myfilt: the former finds
solutions for all problems, the latter only for 15 out of 20
instances.

A possible improvement to \myrepl could be to let it enumerate a
certain amount of solutions to gather more constraints, then add all
these constraints and restart the search.
This is a hybrid approach between \myfilt and \myrepl.
Selecting the right moment to abort enumeration and restart the
solver is crucial to the performance of such a hybrid approach, and
we consider this a worthwhile subject for future investigations.

\section*{Acknowledgments}

This work is partially supported by TUBITAK Grant 111E116. Peter Sch\"{u}ller is supported by TUBITAK 2216 Research Fellowship.

\bibliographystyle{splncs}

\end{document}